\documentclass[letter,11pt]{article}
\usepackage[utf8]{inputenc}
\usepackage{hyperref}
\usepackage{times}
\usepackage{amsmath}
\usepackage[margin=1in]{geometry}
\usepackage{natbib}
\bibpunct{(}{)}{;}{a}{,}{,}
\usepackage{graphicx}

\title{Contextual Word Representations:  \\ A Contextual Introduction}
\author{Noah A. Smith \\
Paul G. Allen School of Computer Science and Engineering, University of Washington \\
Allen Institute for Artificial Intelligence \\
\url{nasmith@cs.washington.edu} 
}

\begin{document}
\maketitle

\begin{abstract}
This introduction aims to tell the story of how we put words into computers.  It is part of the story of the field of natural language processing (NLP), a branch of artificial intelligence.\footnote{For those seeking a textbook about NLP, I recommend \citet{jurafsky-draft} and \citet{eisenstein-18}.}  It targets a wide audience with a basic understanding of computer programming, but avoids a detailed mathematical treatment, and it does not present any algorithms.  It also does not focus on any particular \emph{application} of NLP such as translation, question answering, or information extraction.  The ideas presented here were developed by many researchers over many decades, so the citations are not exhaustive but rather direct the reader to a handful of papers that are, in the author's view, seminal.  After reading this document, you should have a general understanding of word vectors (also known as word embeddings):  why they exist, what problems they solve, where they come from, how they have changed over time, and what some of the open questions about them are.  Readers already familiar with word vectors are advised to skip to Section~\ref{contextual} for the discussion of the most recent advance, contextual word vectors.

The content of this article will appear in ``Contextual Word Representations:   Putting Words into Computers'' in the \emph{Communications of the ACM}, June 2020.
\end{abstract}

\section{Preliminaries} \label{sec:prelim}
There are two ways to talk about words.
\begin{itemize}
    \item A \textbf{word token} is a word observed in a piece of text.  In some languages, identifying the boundaries of the word tokens is a complicated procedure (and speakers of the language may not agree on the ``correct'' rules for splitting text into words), but in English we tend to use whitespace and punctuation to delimit words, and in this document we will assume this problem, known as \emph{tokenization} is ``solved.''  For example, the first sentence of this paragraph is typically tokenized as follows (with the end-of-sentence punctuation treated as a separate token):
    \begin{center}
    A word token is a word observed in a piece of text .
    \end{center}
    There are thirteen tokens in the sentence.
    \item A \textbf{word type} is a distinct word, in the abstract, rather than a specific instance.  Every word token is said to ``belong'' to its type.  In the example above, there are only eleven word types, since the two instances of \emph{word} share the same type, as do the two instances of \emph{a}.  (If we ignored the distinction between upper and lower case letters, then there would only be ten types, since the first word \emph{A} would have the same type as the fifth and ninth words.)    When we count the occurrences of a vocabulary word in a collection of texts (known as a \textbf{corpus}, plural \textbf{corpora}), we are counting the tokens that belong to the same word type.
\end{itemize}

\section{Discrete Words}

In a computer, the simplest representation of a piece of text is a sequence of characters (depending on the encoding, a character might be a single byte or several).  A word type can be represented as a string (ordered list of characters), but comparing whether two strings are identical is costly.

Not long ago, words were usually \emph{integerized}, so that each word type was given a unique (and more or less arbitrary) nonnegative integer value.  This had the advantages that (1) every word type was stored in the same amount of memory, and (2) array-based data structures could be used to index other information by word types (like the string for the word, or a count of its tokens, or a richer data structure containing detailed information about the word's potential meanings).  The vocabulary could be continuously expanded as new word types were encountered (up to the range of the integer data type, over 4 billion for 4-byte unsigned integers).  And, of course, testing whether two integers are identical is very fast.

The integers themselves didn't \emph{mean} anything; the assignment might be arbitrary, alphabetical, or in the order word tokens were observed in a reference text corpus from which the vocabulary was derived (i.e., the type of the first word token observed would get 0, the type of the second word token would get 1 if it was different from the first, and so on).  Two word types with related meanings might be assigned distant integers, and two ``adjacent'' word types in the assignment might have nothing to do with each other.  The use of integers is only a convenience following from the data types available in the fashionable programming languages of the day; in Lisp (for example), ``gensym'' would have served the same purpose, although perhaps less efficiently.  For this reason, we refer to integer-based representations of word types as \textbf{discrete} representations.

\section{Words as Vectors}\label{sec:wav}

To see why NLP practitioners no longer treat word types as discrete, it's useful to consider how words get used in NLP programs.  Here are some examples:
\begin{itemize}
    \item Observing a word token in a given document, use it as evidence to help predict a category for the document.  For example, the word \emph{delightful} appearing in a review of a movie is a cue that the reviewer might have enjoyed the film and given it a positive rating.\footnote{Of course context matters; \emph{the most delightful part of seeing this movie was the popcorn} would signal just the opposite.  See \citet{pang-08} for a detailed treatment of the problems of sentiment and opinion analysis.}
    \item Observing a word token in a given sentence, use it as evidence to predict a word token in the translation of the sentence.  For example, the appearance of the word \emph{cucumber} in an English sentence is a cue that the word \emph{concombre} might appear in the French translation.
    \item Conversely, given the full weight of evidence, choose a word type to write as an output token, in a given context.
\end{itemize}
In each of the above cases, there is a severe shortcoming to discrete word types:  information about how to use a particular word as evidence, or whether to generate a word as an output token, cannot be easily \emph{shared} across words with similar properties.  As a simple example, consider filling in the blank in the following sentence:
\begin{center}
    S.~will eat anything, but V.~hates  \_
\end{center}
Given your knowledge of the world, you are likely inclined to fill in the blank with high confidence as a token of a type like \emph{peas}, \emph{sprouts}, \emph{chicken}, or some other mass or plural noun that denotes food.   These word types \emph{share something} (together with the other words for food), and we'd like for a model that uses words to be able to use that information.\footnote{\label{fn:oov}One situation where this lack of sharing is sorely noticed is in the case of \emph{new words}, sometimes called ``unknown'' or ``out of vocabulary'' (OOV) words.  When an NLP program encounters an OOV word token, say \emph{blicket}, what should it do?  By moving away from discrete words (as we will do in a moment), we've managed to reduce the occurrence of truly OOV word types, by collecting information about an increasingly large set of words \emph{in advance} of building the NLP program.}  To put it another way, our earlier interest in testing whether two words are \emph{identical} was perhaps too strict.  Two non-identical words may be more or less similar.

The idea that words can be more or less similar is critical when we consider that NLP programs, by and large, are built using \textbf{supervised machine learning}, that is, a combination of examples demonstrating the inputs and outputs to a task (at least one of which consists of words) and a mechanism for \emph{generalizing} from those input-output pairings.  Such a mechanism should ideally exploit \emph{similarity}:  anything it discovers about one word should transfer to similar words.

Where might this information about similarity come from?  There are two strands of thought about how to bring such information into programs.  We might trace them back to the rationalist and empiricist traditions in philosophy, though I would argue it's unwise to think of them in opposition to each other.

One strand suggests that humans, especially those trained in the science of human language, \emph{know} this information, and we might design data structures that encode it explicitly, allowing our programs to access it as needed.  An example of such an effort is WordNet \citep{fellbaum-98}, a lexical database that stores words and relationships among them such as synonymy (when two words can mean the same thing) and hyponymy (when one word's meaning is a more specific case of another's).  WordNet also explicitly captures the different \emph{senses} of words that take multiple meanings, such as \emph{fan} (a machine for blowing air, or someone who is supportive of a sports team or celebrity).
Linguistic theories of sentence structure (\textbf{syntax}) offer another way to think about word similarity in the form of categories like ``noun'' and ``verb.''

The other strand suggests that the information resides in artifacts such as text corpora, and we can use a separate set of programs to collect and suitably organize the information for use in NLP.  With the rise of ever-larger text collections on the web, this strand came to dominate, and the programs used to draw information from corpora have progressed through several stages, from count-based statistics, to modeling using more advanced statistical methods, to increasingly powerful tools from machine learning.

From either of these strands (or, more commonly in practice, by intertwining them), we can derive a notion of a word type as a \textbf{vector} instead of an integer.\footnote{A vector is a list, usually a list of numbers, with a known length, which we call its dimensionality.  It is often interpreted and visualized as a direction in a Euclidean space.}  In doing so, we can choose the dimensionality of the vector and allocate different dimensions for different purposes.  For example:
\begin{itemize}
\item Each word type may be given its own dimension, and assigned 1 in that dimension (while all other words get 0 in that dimension).  Using dimensions only in this way, and no other, is essentially equivalent to integerizing the words; it is known as a ``one hot'' representation, because each word type's vector has a single 1 (``hot'') and is otherwise 0.
    \item For a collection of word types that belong to a known class (e.g., days of the week), we can use a dimension that is given binary values.  Word types that are members of the class get assigned 1 in this dimension, and other words get 0.
    \item For word types that are variants of the same underlying root, we can similarly use a dimension to place them in a class.  For example, in this dimension, \emph{know}, \emph{known}, \emph{knew}, and \emph{knows} would all get assigned 1, and words that are not forms of \emph{know} get 0.
    \item More loosely, we can use surface attributes to ``tie together'' word types that look similar; examples include capitalization patterns, lengths, and the presence of a digit.
    \item If word types' meanings can be mapped to magnitudes, we might allocate dimensions to try to capture these.  For example, in a dimension we choose to associate with ``typical weight'' \emph{elephant} might get 12,000 while \emph{cat} might get 9.  Of course, it's not entirely clear what value to give \emph{purple} or \emph{throw} in this dimension.
\end{itemize}

Examples abound in NLP of the allocation of dimensions to vectors representing word types (either syntactic, like  ``verb,'' or semantic, like ``animate''), or to multiword sequences (e.g., \emph{White House} and \emph{hot dog}).  The technical term used for these dimensions is \textbf{features}.  Features can be designed by experts, or they can be derived using automated algorithms.  Note that some features can be calculated even on out-of-vocabulary word types (see footnote~\ref{fn:oov}).  For example, noting the capitalization pattern of characters in an out-of-vocabulary word might help a system guess whether it should be treated like a person's name.

\section{Words as Distributional Vectors:  Context as Meaning} \label{sec:distributional}

An important idea in linguistics  is that words (or expressions) that can be used in similar ways are likely to have related meanings (\citealp{firth-57}; consider our day of the week example above).  In a large corpus, we can collect information about the ways a word type $w$ is used,  for example, by counting the number of times it appears near \emph{every other word type}.  When we begin looking at the full distribution of contexts (nearby words or sequences of words) in a corpus where $w$ is found, we are taking a \emph{distributional} view of word meaning.

\begin{figure*}
    \centering
    \includegraphics[width=0.8\textwidth]{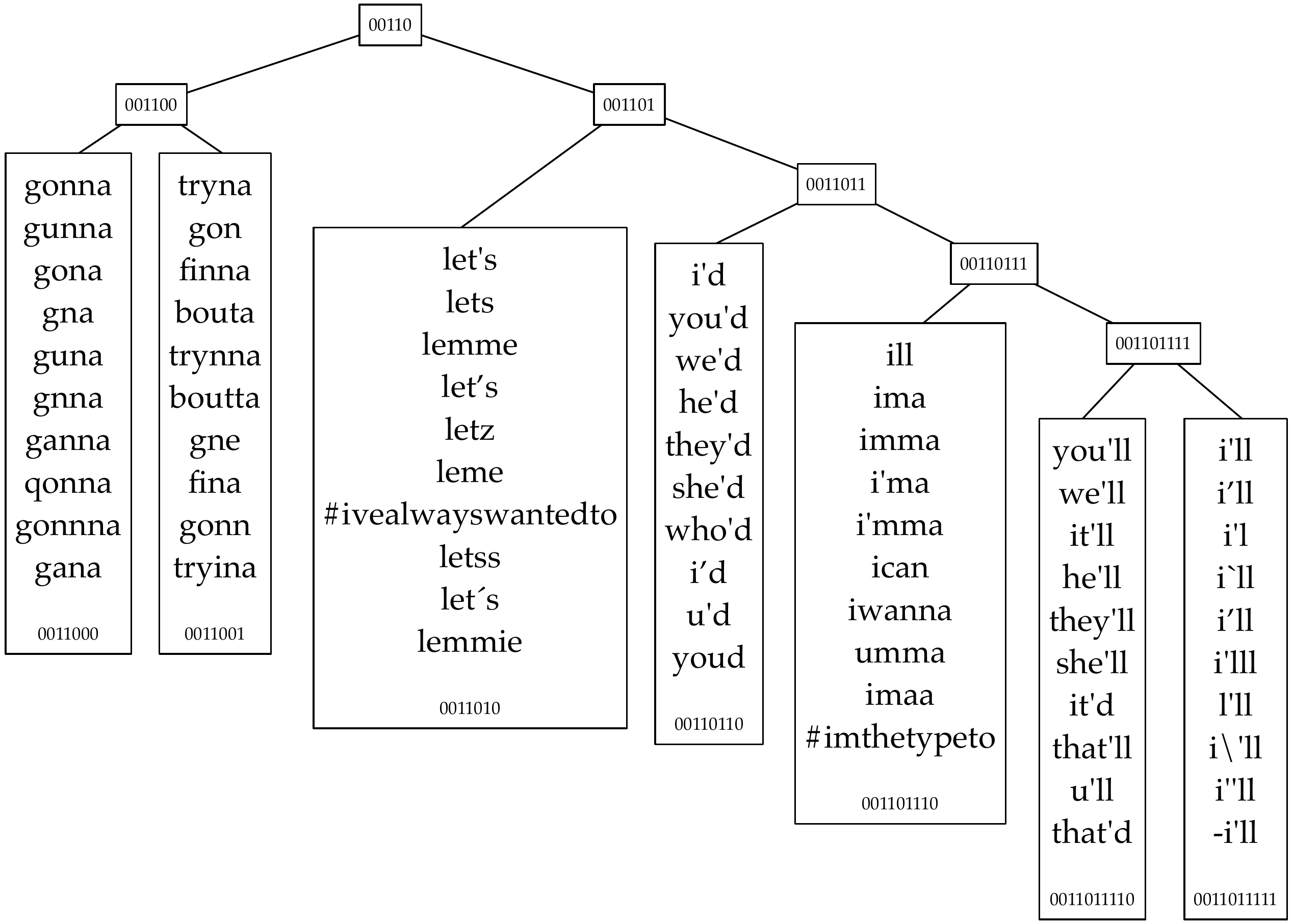}
    \caption{Example Brown clusters.  These were derived from 56M tweets; see \citet{owoputi-13} for details.  Shown are the ten most frequent words in clusters in the section of the hierarchy with prefix bit string 00110.  Intermediate nodes in the tree correspond to clusters that contain all words in their descendants.  Note that differently spelled variants of words tend to cluster together, as do words that express similar meanings, including hashtags. 
    The full set of clusters can be explored at \url{http://www.cs.cmu.edu/~ark/TweetNLP/cluster_viewer.html}. Note that there are several Unicode characters that are visually similar to the apostrophe, resulting in different strings with similar usage.}
    \label{fig:brown}
\end{figure*}

One highly successful approach to automatically deriving features based on this idea is \textbf{clustering}; for example, the \citet{brown-92} clustering algorithm automatically organized words into  clusters based on the contexts they appear in, in a corpus.  Words that tended to occur in the same neighboring contexts (other words) were grouped together into a cluster.  These clusters could then be merged into larger clusters.  The resulting hierarchy, while by no means identical to the expert-crafted data structure in WordNet, was surprisingly interpretable and useful (an example is shown in Figure~\ref{fig:brown}).  It also had the advantage that it could be rebuilt using any given corpus, and every word observed would be included.  Hence suitable word clusters could be built separately for news text, or biomedical articles, or tweets.

\begin{figure}
    \centering
    \small  \begin{tabular}{|c||r|r|r|}
    \hline
    \emph{context words} & $\mathbf{v}(\text{astronomers})$ & $\mathbf{v}(\text{bodies})$ & $\mathbf{v}(\text{objects})$ \\ \hline \hline
't& & & 1 \\ \hline
,& &2 & 1\\ \hline
.& 1 & & 1 \\ \hline
1& & & 1 \\ \hline
And& & &  1\\ \hline
Belt& & &1 \\ \hline
But& 1 & & \\ \hline
Given& & & 1\\ \hline
Kuiper& & &1 \\ \hline
So& 1 & & \\ \hline
and& &1 & \\ \hline
are& & 2&1  \\ \hline
between& & & 1\\ \hline
beyond& &1 & \\ \hline
can& & &1 \\ \hline
contains& &1 & \\ \hline
from& 1 & & \\ \hline
hypothetical& & & 1\\ \hline
ice& &1  & \\ \hline
including& & 1& \\ \hline
is& 1 & & \\ \hline
larger& &1 & \\ \hline
now& 1 & & \\ \hline
of& 1 & & \\ \hline
only& & &1 \\ \hline
out& &1 & \\ \hline
potential& & 1& \\ \hline
the& 1 & &1 \\ \hline
these& &2  &1 \\ \hline
they& 1 & & \\ \hline
think&2  & & \\ \hline
those& & &1 \\ \hline
thought& &2 & \\ \hline
what& 1 & & \\ \hline
\end{tabular}

\begin{equation*}
    \mathrm{cosine\_similarity}(\mathbf{u}, \mathbf{v}) = \frac{\mathbf{u} \cdot \mathbf{v}}{\|\mathbf{u}\| \cdot \|\mathbf{v}\|}
\end{equation*}

\begin{tabular}{|c|rrr|}
\hline
& astronomers & bodies & objects \\ \hline
astronomers & $\displaystyle \frac{14}{\sqrt{14}\cdot \sqrt{14}} = 1$ & $\displaystyle \frac{0}{\sqrt{24}\cdot \sqrt{14}} = 0$ &  $\displaystyle\frac{1+1}{\sqrt{14} \cdot \sqrt{16}} \approx 0.134$\\  & & & \\
bodies & & $\displaystyle \frac{24}{\sqrt{24}\cdot \sqrt{24}}  = 1$ & $\displaystyle\frac{2+2+2}{\sqrt{24}\cdot\sqrt{16}} \approx 0.306$\\  & & & \\
objects & & & $\displaystyle \frac{16}{\sqrt{16}\cdot \sqrt{16}}  = 1$ \\ \hline
\end{tabular}
    \caption{Example calculation of word vectors.  We consider three word types occurring a science news story (\url{https://bit.ly/2B9uaKr}):
    \emph{astronomers}, \emph{bodies}, and \emph{objects}.  The table above shows the frequency of each word occurring within two positions on either side of the word whose vector we are constructing, giving three (vertical) word vectors with 34 visible dimensions (zeroes and other dimensions not shown).   Do you expect \emph{bodies} to be more similar to \emph{astronomers} or to \emph{objects}?  The calculation beneath is the cosine similarity score, applied to each word paired with the others (and itself, always giving similarity of one).  In this tiny corpus, \emph{bodies} is closer to \emph{objects} than either is to \emph{astronomers}.}
    \label{fig:vectors}
\end{figure}


\begin{figure}
    \centering
    \includegraphics[scale=0.8]{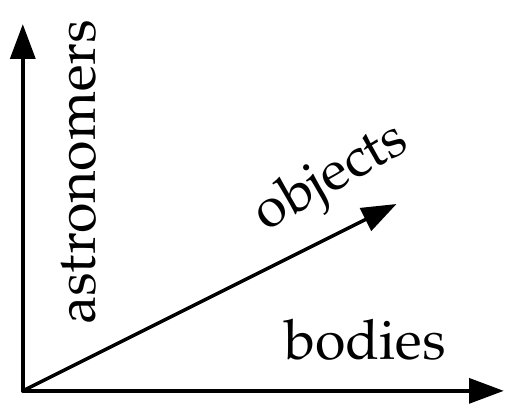}
    \caption{Approximate visualization of the relationships between the three word vectors calculated in Figure~\ref{fig:vectors}.}
    \label{fig:vectors-viz}
\end{figure}

Another line of approaches started by creating word vectors in which each dimension corresponded to the frequency the word type occurred in some context \citep{deerwester-90}.  For instance, one dimension might correspond to \emph{the} and contain the number of times the word occurred within a small window of a token \emph{the}.  Contextual patterns on the left or the right, and of varying distances and lengths, might be included.  The result is a vector perhaps many times longer than the size of the vocabulary, in which each dimension contains a tiny bit of information that may or may not be useful.  An example is shown in Figure~\ref{fig:vectors}.  Using methods from linear algebra, aptly named \emph{dimensionality reduction}, these vectors could be compressed into shorter vectors in which redundancies across dimensions were collapsed.  

These  reduced-dimensionality  vectors had several advantages.  First, the dimensionality could be chosen by the NLP programmer to suit the needs of the program.  More compact vectors might be more efficient to compute with, and might also benefit from the lossiness of the compression, since corpus-specific ``noise'' might fall away.  There's a tradeoff, though; longer, less heavily compressed vectors retain more of the original information in the distributional vectors.  While the individual dimensions of the compressed vectors are not easily interpreted, we can use well-known algorithms to find a word's \emph{nearest neighbors} in the vector space, and these were often found to be semantically related words, as one might hope.

Indeed, these observations gave rise to the idea of \emph{vector space semantics} (see \citealp{turney-10} for a survey), in which arithmetic operations were applied to word vectors to probe what kind of ``meanings'' had been learned.  Famously, analogies like ``\emph{man} is to \emph{woman} as \emph{king} is to \emph{queen}'' led to testing whether $\mathbf{v}(\mbox{\emph{man}}) - \mathbf{v}(\mbox{\emph{woman}}) = \mathbf{v}(\mbox{\emph{king}}) - \mathbf{v}(\mbox{\emph{queen}})$.  Efforts to design word vector algorithms to adhere to such properties followed.

The notable disadvantage of reduced-dimensionality vectors is that the individual dimensions are no longer interpretable features that can be mapped back to intuitive building blocks contributing to the word's meaning.  The word's meaning is \emph{distributed} across the whole vector; for this reason, these vectors are sometimes called \textbf{distributed} representations.\footnote{Though \emph{distributional} information is typically used to build \emph{distributed}  vector representations for word types, the two terms are not to be confused and have orthogonal meanings!}

As corpora grew, scalability became a challenge, because the number of observable contexts grew as well.  Underlying all word vector algorithms is the notion that the value placed in each dimension of each word type's vector is a \emph{parameter} that will be \emph{optimized}, alongside all the other parameters, to best fit the observed patterns of the words in the data.  Since we view these parameters as continuous values, and the notion of ``fitting the data'' can be operationalized as a smooth, continuous objective function, selecting the parameter values is done using iterative algorithms based on gradient descent.
Using tools that had become popular in machine learning, faster methods based on stochastic optimization were developed.  One widely known collection of algorithms is available as the \textbf{word2vec} package \citep{mikolov-13}.  
A common pattern arose in which industry researchers with large corpora and powerful computing infrastructure would construct word vectors using an established (often expensive) iterative method, and then publish the vectors for anyone to use.  

\begin{figure}
\begin{center}
\includegraphics[width=0.8\textwidth]{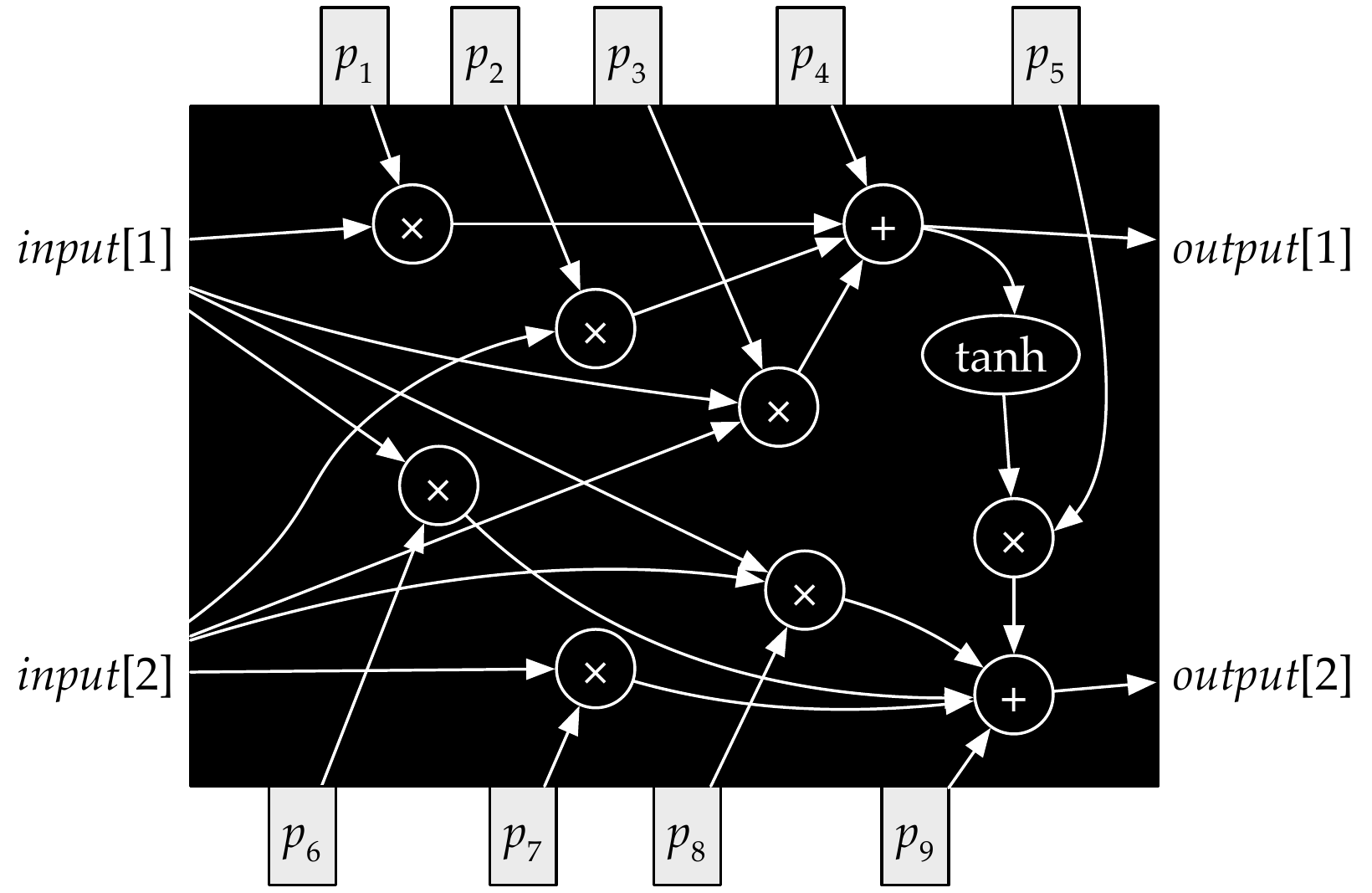}
\end{center}
\caption[]{\label{fig:nn}A neural network is a function from vectors to vectors.  A very simple example is a function from two-dimensional inputs to two-dimensional outputs, such as: 
\begin{minipage}{\linewidth}\begin{align*}
     \mathit{output}[1]& =  &p_1 \times \mathit{input}[1] + p_2 \times \mathit{input}[2] + p_3 \times \mathit{input}[1] \times \mathit{input}[2] + p_4 \\
     \mathit{output}[2]& = &p_5 \times \tanh(\mathit{output}[1]) + p_6 \times \mathit{input}[1] + p_7 \times \mathit{input}[2] \\
      & & +~p_8 \times \mathit{input}[1] \times \mathit{input}[2] + p_9  \\
 \end{align*}  \end{minipage}

 Neural networks are almost always defined in terms of \emph{parameters}, here denoted by $p_1, \ldots, p_9$, which are automatically chosen using standard machine learning algorithms.  Typically, they include at least one transformation that is not linear (e.g., the hyperbolic tangent above).

The illustration displays the toy neural network as a computation graph (inside the black box), with inputs on the left, outputs on the right, and each parameter corresponds to a gray box placed along the top and bottom. Round nodes inside the box correspond to intermediate operations (addition, multiplication, and tanh).
}
\end{figure}

There followed a great deal of exploration of methods for obtaining distributional word vectors.  Some interesting ideas worth noting include:
\begin{itemize}
 \item When we wish to apply neural networks
  to problems in NLP (see Figure~\ref{fig:nn}), it's useful to first map each input word token to its vector, and then ``feed'' the word vectors into the neural network model, which performs a task like translation.  The vectors can be fixed in advance (or \textbf{pretrained} from a corpus, using methods like those above, often executed by someone else), or they can be treated as  parameters of the neural network model, and adapted to the task specifically \citep[e.g.,][]{collobert-11}.  \textbf{Finetuning} refers to initializing the word vectors by pretraining, then adapting them through task-specific learning algorithms.  The word vectors can also be initialized to random values, then estimated solely through task learning, which we might call ``learning from scratch.''\footnote{The result of vectors learned from scratch for an NLP task is a collection of \emph{distributed} representations that were derived from something other than \emph{distributional} contexts (the task data).} 
    \item We can use expert-built data structures like WordNet as additional input to creating word vectors. One approach, \textbf{retrofitting}, starts with word vectors extracted from a corpus, then seeks to automatically adjust them so that word types that are related in WordNet are closer to each other in vector space \citep{faruqui-15}.
    \item We can use bilingual dictionaries to ``align'' the vectors for words in two languages into a single vector space, so that, for example, the vectors for the English word type \emph{cucumber} and the French word type \emph{concombre} have a small Euclidean distance \citep{faruqui-14}.  By constructing a function that reorients all the English vectors into the French space (or vice versa), researchers hoped to align \emph{all} the English and French words, not just the ones in the bilingual dictionary. 
    \item A words' vectors are calculated in part (or in whole) from its character sequence \citep{ling-15}.  These methods tend to make use of neural networks to map arbitrary-length sequences into a fixed-length vector.  This has two interesting effects:  (1) in languages with intricate word formation systems (\textbf{morphology}),\footnote{E.g., the present tense form of the French verb \emph{manger} is \emph{mange}, \emph{manges}, \emph{mangeons}, \emph{mangez}, or \emph{mangent}, depending on whether the subject is singular or plural, and first, second, or third person.} variants of the same underlying root may have similar vectors, and (2) differently-spelled variants of the same word will have similar vectors.  This kind of approach was quite successful for social media texts, where there is rich spelling variation.  For example, these variants of the word \emph{would}, all attested in social media messages, would have similar character-based word vectors because they are spelled similarly:  \emph{would}, \emph{wud}, \emph{wld}, \emph{wuld}, \emph{wouldd}, \emph{woud}, \emph{wudd}, \emph{whould}, \emph{woudl}, and  \emph{w0uld}.  
\end{itemize}

\section{Contextual Word Vectors}\label{contextual}

\begin{figure}
    \centering
    \includegraphics[scale=0.8]{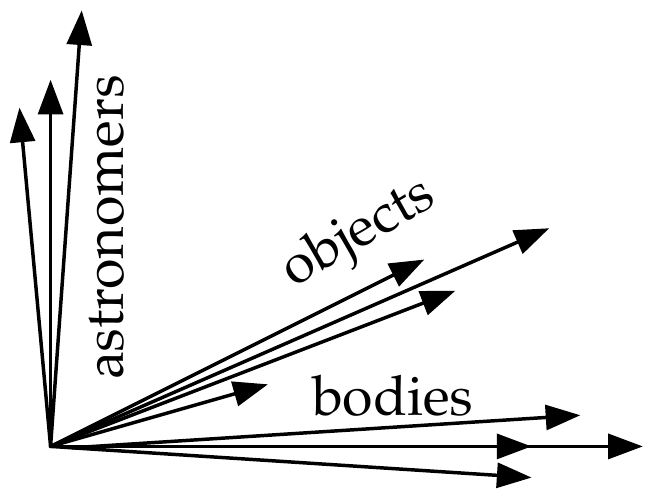}
    \caption{Hypothetical visualization of contextual vectors for tokens of \emph{astronomers}, \emph{bodies}, and \emph{objects} from Figures~\ref{fig:vectors}--\ref{fig:vectors-viz}.}
    \label{fig:vectors-viz2}
\end{figure}

We started this discussion by differentiating between word \emph{tokens} and word \emph{types}.  All along, we've assumed that each word type was going to be represented using a fixed data object (first an integer, then a vector) in our NLP program.  This is convenient, but it makes some assumptions about language that do not fit with reality.  Most importantly, words have different meanings in different contexts.  At a coarse-grained level, this was captured by experts in crafting WordNet, in which, for example, \emph{get} is mapped to  over thirty different meanings (or \textbf{senses}).  It is difficult to obtain widespread agreement on how many senses should be allocated to different words, or on the boundaries between one sense and another; word senses may be \emph{fluid}.\footnote{For example, the word \emph{bank} can refer to the side of a river or to a financial institution.  When used to refer to a blood bank, we can debate whether the second sense is evoked or a third, distinct one.}    Indeed, in many NLP programs based on neural networks, the very first thing that happens is that each word token's type vector is passed into a function that \emph{transforms} it based on the words in its nearby context, giving a new version of the word vector, now specific to the token in its particular context.  In our example sentence from Section~\ref{sec:prelim}, the two instances of \emph{a} will therefore have different vectors, because one occurs between \emph{is} and \emph{word} and the other occurs between \emph{in} and \emph{piece}.  Compare Figures~\ref{fig:vectors-viz} and~\ref{fig:vectors-viz2}.

With hindsight, we can now see that by representing word types independent of context, we were solving a problem that was harder than it needed to be.  Because words mean different things in different contexts, we were requiring that type representations  capture \emph{all} of the possibilities (e.g., the thirty meanings of \emph{get}).  Moving to word token vectors simplifies things, asking the word token representation to capture only what a word means \emph{in this context}.  For the same reasons that the collection of contexts a word type is found in provide clues about its meaning(s), a particular token's context provides clues about its specific meaning.  For instance, you may not know what the word \emph{blicket} means, but if I tell you that I ate a strawberry blicket for dessert, you likely have a good guess.\footnote{Though such examples abound in linguistics, this one is due to Chris Dyer.}  

Returning to the fundamental notion of similarity, we would expect words that are similar to each other to be good substitutes for each other.  For example, what are some good substitutes for the word \emph{gin}?  This question is hard to answer about the word type (WordNet tells us that \emph{gin} can refer to a liquor for drinking, a trap for hunting, a machine for separating seeds from cotton fibers, or a card game), but easy in a given context (e.g., ``I use two parts gin to one part vermouth.'').  Indeed,  \emph{vodka} might even be expected to have a similar contextual word vector if substituted for \emph{gin}.\footnote{The author does not endorse this substitution in actual cocktails.}

\textbf{ELMo}, which stands for ``embeddings from language models'' \citep{peters-18}, brought
a powerful advance in the form of word \emph{token} vectors---i.e., vectors for words in context, or \textbf{contextual word vectors}---that are pretrained on large corpora.    There are two important insights behind ELMo:
\begin{itemize}
    \item If every word token is going to have its own vector, then the vector should depend on an arbitrarily long context of nearby words.  To obtain a ``context vector,'' we start with word type vectors, and pass them through a neural network that can transform arbitrary-length sequences of left- and/or right-context word vectors into a single fixed-length vector.  Unlike word \emph{type} vectors, which are essentially lookup tables, contextual word vectors are built from both type-level vectors and neural network parameters that ``contextualize'' each word.  ELMo trains one neural network for left contexts (going back to the  beginning of the sentence a token appears in) and another neural network for right contexts (up to the end of the sentence).  Longer contexts, beyond sentence boundaries, are in principle possible as well.
    \item Recall that estimating word vectors required ``fitting the data'' (here, a corpus) by solving an optimization problem.  A longstanding data-fitting problem in NLP is \textbf{language modeling}, which refers to predicting the next word given a sequence of ``history'' words (briefly alluded to in our filling-in-the-blank example in Section~\ref{sec:wav}).  Many of the word (type) vector algorithms already in use were based on a notion fixed-size contexts, collected across all instances of the word type in a corpus.  ELMo went farther, using arbitrary-length histories and directly incorporating the language models  known at the time to be most effective (based on recurrent neural networks; \citealp{sundermeyer-12}).  Although recurrent networks were already widely used in NLP (see \citet{goldberg-17} for a thorough introduction), training them as language models, then using the context vectors they provide for each word token as pretrained word (token) vectors was novel.
\end{itemize}

It's interesting to see how the ideas around getting words into computers have come full circle.  The powerful idea that text data can shed light on a word's meaning, by observing the contexts in which a word appears, has led us to try to capture a word \emph{token}'s meaning primarily through the specific context it appears in.  This means that every instance of \emph{plant} will have a different word vector; those with a context that look like a context for references to vegetation are expected to be close to each other, while those that are likely contexts for references to manufacturing centers will cluster elsewhere in vector space.  Returning to the example in Figure~\ref{fig:vectors}, while instances of \emph{bodies} in this article will likely remain closer to \emph{objects} than to \emph{astronomers}, in a  medical news story, we might expect instances of \emph{bodies} to be closer to \emph{humans} (and by extension \emph{astronomers}).

\paragraph{Why is this advance so exciting?}
Whether the development of contextual word vectors completely solves the challenge of ambiguous words remains to be seen.
New ideas in NLP are often tested on benchmark tasks with objectively measurable performance scores.
 ELMo was shown to be extremely beneficial in NLP programs that 
\begin{itemize}
\item answer questions about content in a given paragraph (9\% relative error reduction on the SQuAD benchmark),
\item  label the semantic arguments of verbs (16\% relative error reduction on the Ontonotes semantic role labeling benchmark), 
\item label expressions in text that refer to people, organizations, and other named entities (4\% relative error reduction on the CoNLL 2003 benchmark), and
\item resolve which referring expressions refer to the same entities (10\% relative error reduction on the Ontonotes coreference resolution benchmark).
\end{itemize}
Gains on additional tasks were reported by \citet{peters-18} and later by other researchers.   
\citet{howard-18} introduced a similar approach, ULMFiT, showing a benefit for text classification methods.
A successor approach, bidirectional encoder representations from transformers (BERT; \citealp{devlin-18}) that introduced several innovations to the learning method and learned from more data, achieved a further 45\% error reduction (relative to ELMo) on the first task and 7\% on the second.  On the SWAG benchmark, recently introduced to test grounded commonsense reasoning \citep{zellers-18}, \citet{devlin-18} found that ELMo gave 5\% relative error reduction compared to non-contextual word vectors, and BERT gave another 66\% relative to ELMo.  A stream of papers since this article was conceived have continued to find benefits from 
creative variations on these ideas, resulting in widely adopted models like  GPT-2 \cite{radford-19}, RoBERTa \cite{liu-19b}, T5 \cite{raffel-19}, XLM \cite{lample-19}, and XLNet \cite{yang-19}.
 It is rare to see a single conceptual advance that consistently offers large benefits across so many different NLP tasks.

At this writing, there are many open questions about the relative performance of the different methods.  A full explanation of the differences in the learning algorithms, particularly the neural network architectures, is out of scope for this introduction, but it's fair to say that the space of possible learners for contextual word vectors has not yet been fully explored; see \citet{peters-18b} for some exploration.  Some of the findings on BERT suggest that the role of finetuning may be critical; indeed, earlier work that used pretrained language models to improve text classification \emph{assumed} finetuning was necessary \citep{dai-15}.  While ELMo is derived from language modeling, the modeling problem solved by BERT (i.e., the objective function minimized during estimation) is rather different.\footnote{BERT pretraining focuses on two tasks:  (i) prediction of words given contexts on both sides (rather than one or the other) and (ii) predicting the words in  a sentence given its preceding sentence.}  The effects of the dataset used to learn the language model have not been fully assessed, except for the unsurprising pattern that  larger datasets tend to offer more benefit.

\section{Cautionary Notes}

\paragraph{Word vectors are biased.} Like any engineered artifact, a computer program is likely to reflect the perspective of its builders.  Computer programs that are built from data will reflect what's in the data---in this case, a text corpus.  If the text corpus signals associations between concepts that reflect cultural biases, these associations should be expected to persist in the word vectors and any system that uses them.  Hence it is not surprising that NLP programs that use corpus-derived word vectors associate, for example, \emph{doctor} with male pronouns and \emph{nurse} with female ones.  Methods for detecting, avoiding, and correcting unwanted associations is an active area of research \citep{bolukbasi-16,caliskan-17}.  The advent of contextual word vectors offers some possibility of new ways to avoid unwanted generalization from distributional patterns. 

\paragraph{Language is a lot more than words.}  Effective understanding and production of language is about more than knowing word meanings; it requires knowing how words are put together to form more complicated concepts, propositions, and more.  The above is not nearly the whole story of NLP; there is much more to be said about approaches to dealing with natural language syntax, semantics, and pragmatics, and how we operationalize tasks of understanding and production that humans perform into tasks for which we can attempt to design algorithms.  One of the surprising observations about contextual word vectors is that, when trained on very large corpora, they make it easier to disambiguate sentences through various kinds of syntactic and semantic parsing; it is an open and exciting question how much of the work of understanding can be done at the level of words in context.

\paragraph{Natural language processing is not a single problem.}  While the gains above are quite impressive, it's important to remember that they reflect only a handful of benchmarks that have emerged in the research community.  These benchmarks are, to varying degrees, controversial, and are always subject to debate.  No one who has spent any serious amount of time studying NLP believes they are ``complete'' in any interesting sense.   NLP can only make progress if we have ways of objectively measuring progress, but we also need continued progress on the design of the benchmarks and the scores we use for comparisons.  This aspect of NLP research is broadly known as \textbf{evaluation}, and includes both human-judgment-based and automatic methods.  Anyone who is an enthusiast of NLP (or AI more generally) should take the time to learn how progress is measured and understand the shortcomings of evaluations currently in use.

\section{What's Next}

Over the next few years, I expect to see new findings that apply variations on contextual word vectors to new problems and that explore modifications to the learning methods.  For example, building a system might involve sophisticated protocols in which finetuning and task-specific training are carried out on a series of dataset/task combinations.  Personally, I'm particularly excited about the potential for these approaches to improve NLP performance in settings where relatively little supervision is available.  Perhaps, for example, ELMo-like methods  can improve NLP for low-resource genres and languages \cite{mulcaire-19}.  Likewise, methods that are computationally less expensive have the potential for broad use (e.g., \citet{gururangan-19}).  I also expect there will be many attempts to characterize the generalizations that these methods are learning (and those that they are not learning) in linguistic terms; see for example \citet{goldberg-19} and \citet{liu-19}.

\section{Further Reading}
An introductory guide to linguistics for those interested in NLP is provided by \citet{bender-13} and \citet{bender-19}.
A more thorough mathematical treatment of the topics covered in Sections~\ref{sec:prelim}--\ref{sec:distributional} is given in chapter 14 of \citet{eisenstein-18}.  For contextual word vectors, the original papers are recommended at this writing \citep{peters-18,devlin-18}.

%
\section*{Acknowledgments}
Errors and oversights are the author's alone.  The exposition benefited  from feedback from Oren Etzioni, Chris Dyer, numerous students from the University of Washington's winter 2019 offering of CSE 447, and three anonymous reviewers, which is acknowledged with gratitude.  NSF grant IIS-1562364 supported the author's research related to this article.

%
 \bibliographystyle{plainnat}
\bibliography{sample-base}

\end{document}